\documentclass[acmsmall,nonacm]{acmart}
\AtBeginDocument{%
  }

\setcopyright{cc}
\setcctype{by}
\acmJournal{TKDD}
\acmYear{2026} \acmVolume{1} \acmNumber{1} \acmArticle{}
\acmMonth{1} \acmDOI{10.1145/3863957}

\newcommand{\sectionquote}[2]{%
  \begin{center}
    \small\itshape ``#1''\\[2pt]
    \normalfont\small--- #2
  \end{center}\vspace{0.5em}%
}

\usepackage{xcolor}

\usepackage{microtype}
\usepackage{enumitem}
\begin{document}

\title{Not All Interpretation Methods Are Created Equal: Tiered Evaluation of AI Interpretability}

\author{Nicolas Sanchez}
\email{nicolas\_sanchez\_ep@berkeley.edu}
\author{Bin Yu}
\orcid{0000-0002-8888-4060}
\email{binyu@berkeley.edu}
\correspondingauthor
\affiliation{%
  \institution{University of California, Berkeley}
  \city{Berkeley}
  \state{CA}
  \country{USA}
}


\begin{abstract}As AI transitions to sociotechnical deployments, uninterpretable models pose serious threats to reliability, public trust and contestability. However, not all interpretation methods are created equal: for interpretability to enhance deployment safety and effectiveness, interpretation methods must be evaluated in a tiered system to differentiate evidence levels of trustworthiness, rather than validated through isolated interpretation instances.

Our core vision is that elevating the evaluation of interpretation methods to a core AI assessment metric will drive robust knowledge discovery from data (KDD) and ensure safe, contestable sociotechnical deployments of AI. To this end, we propose TRUST, a tiered evaluation framework that builds upon the Predictability-Computability-Stability (PCS) principles for Veridical Data Science (VDS) \cite{yu2020veridical,rewolinski2025pcs,yu2024veridical, yu2013stability} and the Predictive, Descriptive and Relevant (PDR) \cite{murdoch2019definitions} foundations. To differentiate interpretation methods effectively and systematically, TRUST grades interpretation across five tiers: Task-Aligned PDR Foundations, Rigorous Epistemic Grounding, Unifying Semantic Understanding, Scope Transferability, and Targeted Prospective Validation. We present a start-to-finish case study of the iterative Random Forest (iRF) family under TRUST and contrast it with other LLM interpretation methods that struggle to clear specific tiers. Finally, we conclude with brief initial thoughts on how TRUST principles can extend in autonomous agentic frameworks. Ultimately, the TRUST framework fosters contestability and epistemic grounding as foundational evaluation principles for human-machine workflows and, eventually, autonomous agentic workflows.
\end{abstract}




\maketitle

\section{Introduction}
Artificial Intelligence (AI) models are rapidly transitioning from closed-loop novelties into embedded sociotechnical systems. From NHS England providing over half a million clinicians with AI tools \cite{nhsengland2026copilot}, to US Census data showing 37\% of large firms and 55\% of workers using AI \cite{census2026htops, census2026btos}, to expanding classroom partnerships \cite{chatGPTClassroom}, AI is deeply integrated into daily workflows. However, rapid adoption has triggered unintended yet severe vulnerabilities driven by user complacency and automation bias: clinicians override their own accurate medical judgment \cite{tschandl2020human}, lawyers unwittingly submit fabricated precedents \cite{mata2023avianca}, and educators wrongly accuse non-native speaking students of misconduct \cite{liang2023gpt}. These breakdowns have sparked significant reactions, including Norway banning AI from elementary schools \cite{norwayban}, California establishing state AI safety verification frameworks \cite{californiaAIpolicy}, and congressional investigations into UnitedHealth’s nH Predict for improperly denying care \cite{nHPredict}. Ultimately, these examples reveal a systemic flaw: current evaluation frameworks rely on out-of-context benchmarks that ignore deployment realities. While paradigms like Green Shielding \cite{li2026green} evaluate outputs under everyday variations and propose mitigation strategies to address failure modes, AI systems must go further by satisfying the foundational societal principle of contestability \cite{vaccaro2019contestability}.

Contestability has historically referred to mechanisms for users to understand, construct, shape and challenge model predictions or decisions \cite{mulligan2019shaping}. These mechanisms establish accountability and empower steerability in socially deployed systems. Bamberger, Mulligan and Wang \cite{bamberger2026contestability} have demonstrated the need to expand this definition beyond an ex-post feature: contestability must be a systemic property. AI systems must possess reflexive architectures, designed to surface embedded value choices and provide structured pathways to challenge model reasoning not just at the output level, but at configuration and design time. Trustworthy AI model interpretability offers an indispensable avenue for addressing these requirements in addition to guiding trustworthy AI model development and ensuring they enable rigorous knowledge discovery from data. However, not all AI interpretation methods are created equal. Fragile, unreliable interpretation methods are counter-productive for both knowledge discovery and contestability: by providing erroneous perceptions of AI model reasoning, the false sense of security deepens societal mistrust in AI systems. 

To better evaluate trustworthy interpretation methods for trustworthy deployment and contestability requirements, we propose a five-tiered evaluation framework designed to grade interpretation methods for fidelity, safety and utility. Each tier demands progressively stronger independent evidence starting from baseline requirements to gold-standard external validation: Task-Aligned Predictive, Descriptive, and Relevant (PDR) Foundations; Rigorous Epistemic Grounding; Unifying Semantic Understanding; Scope Transferability; and Targeted Prospective Validation (TRUST). In conjunction with the proposed framework, we also provide a standardized TRUST Card \footnote{An interactive form with the TRUST card is available at \url{https://forms.gle/jro7XjNhWvi6wDD18}} designed both for documented evaluation and guided development. The TRUST tiers build upon two established pillars towards trustworthy and safe AI including contestability. First, it directly expands the foundational definitions of interpretable machine learning set forth by Murdoch et al. \cite{murdoch2019definitions}. Second, it relies on the principles of Veridical Data Science (VDS) \cite{yu2020veridical,rewolinski2025pcs,yu2024veridical,yu2013stability}.

Within Veridical Data Science (VDS), the core principles of Predictability, Computability, and Stability (PCS) examine reasonable human judgment calls and aggregate good or reality-checked ones across every stage of the AI pipeline \cite{yu2020veridical}. Grounded in PCS, the TRUST framework assesses the trustworthiness of interpretation methods throughout the AI lifecycle—from initial design, training and pre-deployment validation to live monitoring and post-mortem auditing. To accommodate diverse stakeholders with varying domain and technical expertise, TRUST provides a multi-dimensional, tiered evaluation framework where later tiers demand progressively stronger external independent evidence.

For the remainder of this paper, we detail the five tiers of TRUST, followed by a start-to-finish demonstration using the iterative Random Forest (iRF) in biological settings as well as other LLM interpretation methods (CD-T, SPEX, GCT). In our concluding discussion, we share initial thoughts on extending TRUST to autonomous multi-agent systems.

\begin{figure}[t]
  \centering
  \includegraphics[width=0.6\columnwidth]{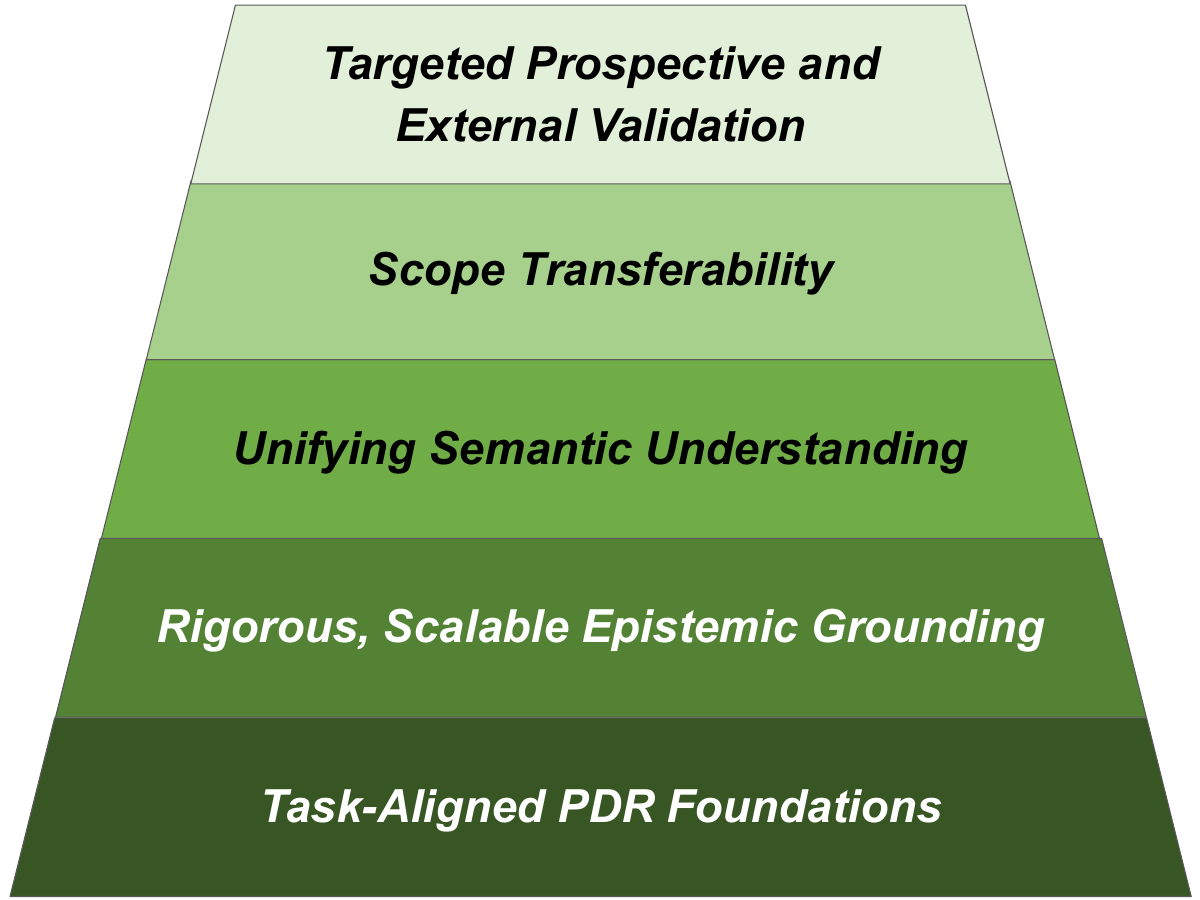}
  \caption{Bottom-Up TRUST Evaluation Framework}
  \Description{A diagram showing the five evaluation tiers starting from task-aligned foundations up to prospective validation.}
  \label{fig:trust_framework}
\end{figure}

\section{Bottom-Up TRUST Evaluation Framework}
\subsection{Tier 1 : Task-Aligned Predictive, Descriptive, and Relevant (PDR) Foundations}
\sectionquote{There's no sense in being precise when you don't even know what you're talking about.}{John von Neumann (Method in the Physical Sciences)}

The first and most basic tier directly builds on the formalization of interpretability introduced by Murdoch et al. \cite{murdoch2019definitions}.  An interpretation method, and its associated AI system must, as a baseline, satisfy three desiderata to ensure it provides both reliable and sociotechnically actionable insights:
\paragraph{Reality Check (“P”)} The scope of predictive accuracy broadens to encompass the strength of the AI system’s reality checks, ensuring that the interpreted system captures true underlying relationships in the data. Notably, accuracy should not be restricted to standard average accuracy. Instead, evaluation requires multiple metrics for accuracy  (e.g. sensitivity, specificity, worst-case error, top-K) stratified across key subpopulations that are tailored for the task.
\paragraph{Descriptive Accuracy (“D”)} An interpretation method’s outputs must faithfully reflect the AI system’s internal decision-making logic. This includes the need for outputs to be cleanly translatable and understandable to the desired stakeholders without sacrificing fidelity.
\paragraph{Relevance (“R”)} This demands that the communicated information proves useful to specific target stakeholders for a specific task. Crucially, the choice of task and target stakeholders is directly tied to the purpose assigned to the AI system being interpreted, and the human workflow it is to be embedded within. This is especially important for contestability: a descriptive explanation that fails to incorporate normative foundations becomes irrelevant for contestation since it falls short of empowering subjects to meaningfully challenge or affect the underlying value alignment of the system.

\subsection{Tier 2 : Rigorous and Scalable Epistemic Grounding}
\sectionquote{[T]he user of anything is the one who knows most of it by experience[...] . As, for example, the flute-player reports to the flute-maker which flutes respond and serve rightly in flute-playing, and will order the kind that must be made, and the other will obey and serve him.}{Plato’s Republic Book X}
Trusted incumbent domain experts provide a uniquely critical reality check (“P” in PCS). They both possess unique contextual knowledge of domain-specific dynamics and are responsible for ensuring that the AI system does not corrupt the workflow’s purpose upon integration \cite{mulligan2019shaping}. However, with the growth of data availability, AI outputs have scaled with more compute at a much faster pace than human expert bandwidth. Exhaustive, instance-by-instance review thus becomes unviable. This scaling bottleneck directly connects to the computability principle (“C” in PCS) that here extends from algorithmic computation to encompass resource-realistic feasibility in terms of human time and cognitive effort. Tier 2, Rigorous and Scalable Epistemic Grounding, demands that interpretation methods explicitly identify this trustworthy expert and structure their outputs to enable systematic, efficient review at scale. Three criteria are of notable importance for addressing Tier 2.

\paragraph{Objective Expert Validation} The expert or other reference that will serve to review the interpretation method outputs must be independent, established and trusted sources of domain knowledge. The presentation of the output must actively mitigate confirmation bias and automation complacency during review. Interpretation methods often risk being persuasive by presenting plausible-seeming rationalizations for system decisions. To fight this, review procedures should emphasize controlled methodologies, including blinded expert evaluations and presentation of negative controls \cite{gaube2021ai, adebayo2018sanity}.

\paragraph{Cognitively Compatible Representations} The interpretation method output presentation should strive to be compatible with established domain terminology or format (e.g. mathematical equations, logical statements or standardized rubrics). By presenting outputs in a familiar, easy-to-reference format, limited expert cognitive resources can focus on critical decision logic rather than parsing novel  presentation. However, this push for compatibility should not come at the cost of precluding an interpretation method from presenting a novel form of insight that challenges existing paradigms.

\paragraph{Scalable Auditing Process} The selection of data or examples for expert review must intentionally balance realistic expert bandwidth with rigor, directly aligning with the sociotechnical purpose and priorities of the system (e.g. outputs close to the model decision boundary or within high-risk subpopulations). Furthermore, clustering techniques can group recurring logic and enable structural rather than isolated instance level review. When tested for safety and trustworthiness, specialized expert AI systems can also be integrated to perform semi-automated auditing. \\

\subsection{Tier 3 : Unifying Semantic Understanding}
\sectionquote{If names are not correct, language will not be in accordance with the truth of things.}{Confucius (The Analects)}
In line with the stability principle (“S”) of PCS, Tier 3, Unifying Semantic Understanding among Stakeholders, evaluates whether an interpretation method’s outputs provide unambiguous and stable insights across relevant stakeholders. 

\paragraph{Baseline Output Stability}  Fundamentally, unified understanding requires stability and consistency in the interpretation method’s outputs against reasonable perturbations within the context of the task. Similar inputs processed through the same internal model (within reasonable data perturbations or random noise) must yield functionally similar interpretation outputs. Otherwise, a fragile interpretation method will inevitably lead to conflicting conclusions.

\paragraph{Linguistic Stability} Jargon collision between disciplines is of particular concern. The word “matrix” may refer to tabular data for an engineer but relate to cellular micro-environments for a biologist. Similarly, “bias” may mean a systematic deviation of an estimator for a statistician versus suggesting issues of fairness and discrimination to a regulator. Interpretation methods must communicate the underlying decision-process using shared and stable language that reduces the likelihood of conflicting assumptions.

\paragraph{Representative Examples} The unified semantic understanding should be validated across intentionally chosen and clearly documented representative examples. Following the Population, Question of interest, Representativeness and Scrutiny (PQRS) framework \cite{yu2018artificial}, representativeness assesses how closely the chosen examples relate to the target population of potential deployment-usages . The statistical definition of population is the complete set of conditions, environments, and subpopulations under which the data and the output are generated in the real-world \cite{yu2018artificial}. Notably, the usage of the interpretation method may differ slightly from the target population of the underlying deployed AI system. Interpretation is often disproportionately invoked in high-consequence or edge-case decisions where agreement from different stakeholders is most critical. Formal sampling strategies such as stratified random sampling across domain-critical subgroups, or decision boundary regions should be carefully documented (e.g. sampling method description or code, inclusion/exclusion criteria) to convincingly justify that these chosen examples achieve target population representativeness.  If random sampling is not used, as often is the case for benchmarks, documentation is needed to make the case to describe the target population for the benchmark and how it achieves representativeness.\\

\subsection{Tier 4 : Scope Transferability}
\sectionquote{To foresee consists in seeking to apply to a new case what we have observed in a former case; but the new case will never be identical with the former.}{Henri Poincaré (Science and Hypothesis)}

The stability principle (“S”) also requires that interpretation methods intended for real-world deployment remain robust against new sources of datasets and related or even far-away domains. At deployment, the AI system will be exposed to a variety of deployment conditions and, as such, distinct target populations. Data curators may use different labeling standards or record data using different hardware. Engineers may store and pre-process that data through different pipelines. Domain practitioners may employ differing protocols for how and when the AI system is used. As a result, it is critical to understand if the interpretation methods remain trustworthy under these changes. Tier 4, Scope Transferability, therefore evaluates an interpretation method’s proven capacity to generalize beyond its original dataset and context, to a broader class of datasets or related or even far-away tasks. 

\paragraph{Broad Validation} Historically, supervised learning models have been rigorously vetted against at least dozens of benchmark datasets if not hundreds to prove their efficacy. In contrast, interpretability methods are frequently validated on only a handful of highly specific examples. Tier 4 instead demands empirical validation across a broad spectrum of public and private datasets to match the scale of early supervised learning by testing across at least dozens of diverse tasks.

\paragraph{Demarcation of Applicability} No single interpretation method will work for all possible domains. Nevertheless, testing across a variety of datasets provides evidence to clearly define the areas of applicability of the method.

\paragraph{Controlled Setting Validation} One notable form of scope transferability covered under Tier 4 involves testing interpretation methods under controlled conditions with well-understood ground truth. Demonstrating the success of methods in recovering those ground-truth dynamics in realistic synthetic settings or simplified but still realistic versions of the task provides evidence of the interpretation method’s performance before justifying its applicability in more open-ended, real world environments.

\subsection{Tier 5 : Targeted Prospective and External Validation
}
\sectionquote{Experience does not err. Only your judgments err by expecting from her what is not in her power.}{Leonardo Da Vinci (Codex Atlanticus)}
Tier 5, Targeted Prospective and External Validation, demands what is, at the moment, arguably the ultimate end-goal of interpretation methods: to generate insights that lead stakeholders to discover new, actionable knowledge verifiable through targeted prospective and external validation. While Tier 4 touches on this by focusing on robustness on retrospective datasets, Tier 5 goes further by going from observational evaluation to prospective real-world execution.

\paragraph{Actionable Interventions}  To clear Tier 5, interpretation method outputs must surface concrete testable targets for prospective intervention. This could include novel targeted data collection, prospective wet-lab experiments or randomized intervention field trials. 
\paragraph{Empirical Confirmation}Tier 5 further demands that these actions result in empirically confirmed positive outcomes, defined in close relation to the purpose of the underlying AI system and the relevant stakeholders. For scientific discovery, an interpretation method may lead to a high-replicability rate experiment that justifies a novel scientific law or principle. For healthcare or public policy, the interpretation method may lead to intervention that improves health outcomes or quality of life. The interpretation method should make clear what the positive outcome is and who it benefits with clear documentation.
\paragraph{Deployment Endorsement} The transition from pure algorithmic insight to real-world experimentation carries a strong implicit endorsement for deployment: to suggest and execute a relevant physical experiment requires a deeper understanding of operational dynamics and feasibility. Clearing Tier 5 establishes the successful execution of intentional, external and prospective validation as the strongest proof of an interpretation’s actionable relevance and real-world stability within a purpose-driven system.

\section{A Gold Standard in Biology: Iterative Random Forest (iRF)}
The TRUST framework is broadly applicable across a wide class of AI systems. However, the core principles were inspired by the development of iterative Random Forest (iRF) \cite{basu2018iterative,wang2025epistasis, kumbier2018signed}, an algorithm (and its variants) that adds stability to random forests to enable  interpretable interactions.  Evaluating iRF serves as a start-to-finish demonstration of TRUST.

iRF operates under a binary classification setup where stability is added to random forests (RF) through weighted sampling of the features for splits based on MDI importance scores from the last iteration and the first iteration uses RF. Two features that fall on the same path of a tree in the forest are interpreted as interacting. Random Intersection Trees (RIT) traverse the paths and extract candidate Boolean feature
interactions.  Interactions are assigned stability scores via outer-layer bootstrapping, and only stable ones are preserved. 

The base iRF identifies importances from single features and Boolean feature interactions. Signed iRF (siRF) incorporates the directional nature of the feature splits while low-signal iRF (lo-siRF) adapts the pipeline to low-signal settings. We consider iRF across three biological settings:

\begin{description}
    \item \textit{Drosophila Embryogenesis} iRF predicts enhancer status based on transcription factor bindings, histone modifications and chromatin accessibility to guide  developmental biologists \cite{basu2018iterative,kumbier2018signed}
    \item \textit{Alternative Splicing in Human Cells} iRF predicts RNA splicing rates from histone modifications and binding proteins to help biologists uncover co-transcriptional regulatory mechanisms \cite{basu2018iterative}
    \item \textit{Hypertrophic Cardiomyopathy (HCM)} iRF predicts left ventricular mass (LVM) to help cardiovascular biologists discover gene and multi-gene epistatic interactions responsible for a specific heart disease (HCM)  \cite{wang2025epistasis}.
\end{description}

\subsection{Tier 1: Task-Aligned PDR Foundations
}
iRF satisfies the PDR Foundations across all three biological settings.

\textit{"P"} In Drosophila, iRF matched state-of-the-art standard Random Forests.  In human splicing, iRF increased held-out test set performance over random forests (from 0.44 to 0.51 AUC-PR) while concentrating tree decision paths on fewer features. In HCM, a lower signal regime due to the complex human disease, lo-siRF, slightly surpassed other baselines in classification accuracy while utilizing Boolean interactions much closer to biologically backed epistatic interactions \cite{wang2025epistasis}.

\textit{"D"} The outer-loop bootstrap mechanism identifies stable Boolean decision paths that directly reflect the internal mechanics of the predictive engine.

\textit{"R"} Across all three studies, predicted outcomes (enhancer activity, splicing rates, LVM phenotype) are relevant biological targets. Similarly, inputs and interpretation method outputs (epigenetic markers, single-nucleotide variants) align with accepted biological mechanical drivers of each outcome.  Decision tree thresholding also more faithfully reflects biological expression thresholds in biomolecular interaction processes.

\subsection{Tier 2: Rigorous and Scalable Epistemic Grounding}
iRF also addresses Tier 2 through intentional interactive review by experts or by comparing against established domain knowledge.

\textit{Rigor} In Drosophila, 20 second-order interactions uncovered by iRF were compared against established literature. In the HCM study, uncovered gene and gene-gene interactions were reviewed by cardiovascular biologists in a blinded manner with negative controls (mid-ranked interactions and random interactions).

\textit{Scalability} RIT+bootstrap displays algorithmic scalability, successfully navigating massive search space (> $10^9$ candidates for HCM). Human review is facilitated by outputting ranked, highly stable Boolean rules that are easily understandable to biologists. However, exceedingly large candidate pools still challenge true scalability of the biologist review.

\subsection{Tier 3: Unifying Semantic Understanding (Among Stakeholders)
}
iRF robustly satisfies Tier 3. iRF and RIT+bootstrap both emphasize high stability features and outputs, ensuring that findings will be consistent across runs and hence reproducible. Further, explicit Boolean logic with domain-meaningful variables makes for clear logic-based findings that provide unambiguous meaning across biologists.

\subsection{Tier 4: Scope Transferability}
iRF also addresses scope transferability though extension to more datasets and domains is warranted. As demonstrated by the three studies, iRF architecture shows generalization across different biological domains, from fruit flies to human cardiac disease \cite{wang2025epistasis, basu2018iterative}. iRF also displays performance in simplified settings. In Drosophila, the framework is shown to be robust when enhancer activity follows Local Spiky Sparse (LSS) dynamics \cite{behr2022provable}. lo-siRF also displayed controlled Type 1 error in purely additive genetic simulations and higher statistical power in epistatic simulations (non-linear gene-gene interactions) compared to linear models \cite{wang2025epistasis,tang2023statistical}. 

\subsection{Tier 5: Targeted Prospective and External Validation}
iRF finally robustly addresses Tier 5. For Drosophila, 80\% of the 20 prioritized order-2 interactions were validated by experiments in the literature, and order-3 and order-4 interactions generated specific, testable hypotheses regarding enhancer activity \cite{basu2018iterative}.  For HCM, 80\% (4 out of 5) of lo-siRF suggested gene and gene interactions were subsequently confirmed via physical wet-lab experiments \cite{wang2025epistasis,tang2023statistical}.

\begin{table*}[t]
\caption{TRUST Evaluation Summary Across Case Studies}
\label{tab:trust_summary}
\small
\centering
\begin{tabular}{lccccc}
\toprule
\textbf{Method} & \textbf{Tier 1} & \textbf{Tier 2} & \textbf{Tier 3} & \textbf{Tier 4} & \textbf{Tier 5} \\
\midrule
\textbf{iRF } & Robust & Partial & Robust & Partial & Robust \\
\textbf{CD-T}          & Partial & Partial & Partial & Partial & Unaddressed \\
\textbf{SPEX} & Partial & Partial & Unaddressed & Partial & Unaddressed \\
\textbf{GCT}  & Partial & Partial & Partial & Partial & Robust \\
\bottomrule
\end{tabular}
\end{table*}

\section{Other Interpretation Method Case Studies
}

To further illustrate the TRUST framework, we evaluate additional interpretation methods. Rather than exhaustive evaluations, these are brief demonstrations to other interpretation paradigms. 

\subsection{Contextual Decomposition for Transformers (CD-T)
}
Contextual Decomposition for Transformers (CD-T) is an internal mechanistic interpretation method for Transformers that captures contributions of combinations of input features or internal components. Guided by PCS, CD-T was used to analyze a BERT model fine-tuned on pathology reports to classify primary Gleason scores, a prognostic cancer score \cite{hsu2025efficient}. More specifically,  CD-T extracts a stable circuit of attention heads from a collection of pathology reports. CD-T then finds influential words by identifying the most highly attended words when averaged across attention maps from that stable circuit. The purpose is to help urologists understand the model reasoning that drives a Gleason score classification. Main stakeholders include the urologist reviewing the BERT prediction seeking clinical rationale, patients whose treatment pathway depends on the cancer score, and healthcare administrators overseeing clinical compliance.

From the perspective of TRUST tiers, CD-T partially addresses Tiers 1 through 3. For Tier 1, the underlying BERT model achieves high accuracy on classification for a well-established prostate cancer score, supporting “R” and "P" in PDR. CD-T also offers higher descriptive faithfulness ("D") than path-patching (Tier 1). CD-T outputs were reviewed by expert urologists, showing text-level attributions highlighted diagnostically relevant criteria for Gleason (Tier 2) and helped them differentiate between more or less performing models. CD-T also supports output stability by being grounded in PCS (Tier 3).  However, one semantic gap negatively affects all three tiers: attention-head circuits represent computation flow rather than clinical diagnostic rationale, rendering them ambiguous to non-technical stakeholders. Without clear evidence that clinical experts can consistently map between the two concepts, both the rigor of the expert review and the unified semantic understanding brought about by CD-T are put into question.

For Tier 4, CD-T possesses strong methodological transferability, generalizing across transformer architectures and tested on non-clinical text (e.g. SST-2, AGNews \cite{hsu2025efficient,hsu2024mechanistic}). However, it fails to compel or produce actionable recourse or concrete safety guards based on these outputs, falling short on Tier 5.

\subsection{Spectral Explainer (SPEX)}

Spectral Explainer (SPEX) focuses on the efficient discovery of feature interactions, notably in black box LLMs \cite{kang2025spex}. By assuming that interactions are sparse, SPEX makes use of Fourier decomposition using sparse graph decoding tools to efficiently find dominant interactions through feature masking. To reduce the number of LLM calls, ProxySPEX \cite{butler2026proxyspex} extends SPEX by using a surrogate gradient-boosted tree (GBT) model to take advantage of the hierarchical structure of interactions and achieve a similar level of faithfulness with one tenth of the expensive LLM calls (emphasizing the "C" in PCS). We consider here the application of SPEX on the HotPotQA dataset \cite{yang2018hotpotqa}, a dataset which requires synthesizing facts across different documents to provide a correct answer. The purpose here is to demonstrate the LLM is in fact incorporating and reasoning from these different facts. The stakeholders would be natural language processing (NLP) and AI safety researchers seeking proof of correct LLM reasoning.

Similar to CD-T, SPEX partially addresses Tiers 1-2 in its construction and evaluation but is weakened by a persistent semantic gap. For Tier 1, the evaluated LLMs achieve high benchmark exact match and F1 score on the HotPotQA, ensuring “P” in PDR. Evaluating feature interactions across documents directly addresses the multi-hop reasoning that is being tested ("R"). Algorithmic descriptive accuracy ("D") is verified by achieving better faithfulness (20\% higher than marginal independent attribution methods). However, complex boolean interactions between words or phrases remain difficult to interpret for researchers and do not directly align with existing knowledge sources. Indeed, for Tier 2, the epistemic grounding relied on comparing overlaps between SPEX outputs and supporting facts as annotated by humans. Human annotations were collected as a flat set of words, questioning the cognitive compatibility with the boolean interaction outputs of SPEX. This ambiguity combined with lack of interactive review further weakens evidence that these outputs will support a unified semantic understanding among stakeholders (Tier 3). 

By virtue of being a model-agnostic, black-box interpretation method, SPEX stands out as highly transferable from a methodological perspective, but is tested only on a handful of datasets. Though SPEX provides audit trails to help detect correct or erroneous reasoning, it  stops short of proving actions or actionable hypotheses to affect or improve the model reasoning, falling short on Tiers 4 and 5.

\subsection{Generative Causal Testing
}
PCS-guided Generative Causal Testing \cite{antonello2024generative} (GCT) provides a striking example of integrating Tier 5 into interpretation method design. Within the field of neuroscience, GCT translates complex brain-encoding models into human-readable hypotheses regarding what specific concepts activate certain brain regions.

GCT trains a predictive voxel-wise encoding model mapping LLM embeddings to fMRI BOLD responses (to capture “P” in PDR and PCS). The model is then run across a large corpus, using contrastive prompts and bootstrapping to find the $n$-grams that consistently elicit the strongest response (“S” in PCS). GCT then outputs an LLM-guided summarization of these $n$-grams as a candidate hypothesis for the specific voxel or region. Using these outputs as blueprints, an LLM then generates a synthetic, coherent story engineered to trigger the targeted voxel to confirm or refute the hypothesized concepts. Primary stakeholders here include neuroscientists evaluating brain-language selectivity.

Starting with Tier 1, GCT uses fMRI BOLD signals, in line with cognitive neuroscience practice ("R") and initially filters for voxels with high predictive power on held-out test data (“P” from PDR).  However, descriptive accuracy, “D” from PDR, is hurt by the LLM inter-mediation. Though a language based summary is clearer to understand for experts (supporting Tiers 2 and 3), the summarized concept can suffer from oversimplification or hallucination by the LLM, severing the direct mechanistic link to the underlying model logic. In terms of epistemic grounding (Tier 2), GCT independently recovered established functional selectivity from previously published neuroscientific literature (such as “Body parts” in the extrastriate body area or “Scenes and settings” in the parahippocampal place area) but lacked an interactive, targeted and blinded review process.

In terms of scope transferability (Tier 4), GCT was shown to work across different scales, from individual voxels to broader cortical
regions of interest, including newly found micro regions in the prefrontal cortex but was not run on a large variety of datasets. Most notably, GCT stands out for its incorporation of Tier 5 in its design. GCT explicitly designs an in vivo experiment to confirm or refute the output from the interpretation methods. Candidate hypotheses serve as a blueprint to design LLM-assisted synthetic stories structured to validate the hypotheses. The resulting physical brain responses directly confirm or refute the causal validity of the interpretation method output. Success rates of these in vivo experiments varied and were overall not as high as the 80\% success rate of iRF.

\section{Discussion}
We present TRUST as a principled, multi-tiered AI evaluation framework. Our core vision is that elevating AI interpretation methods into a standalone evaluation metric is essential for modern AI and KDD. By satisfying the foundational principle of contestability for AI governance while firmly grounding data insights in domain knowledge and actionable interventions, TRUST enables both scientifically rigorous and trustworthy, contestable AI deployment.

Rigorous evaluation of interpretation methods enables human-machine partnerships. In our view, the ideal integration of AI into workflows mirrors the historical relationship between mathematics and intuition. Humans provide intuition and creative direction grounded by their physical experience. AI systems provide the computational engine to validate, refute or refine those intuitions. Rather than working in parallel, the iteration between human intuition and data-driven AI validation can unlock vast new knowledge, provided trustworthy interpretability serves as the critical bridge.
However, elevating interpretability is not intended to restrict AI learning and development to the limits of human cognition. A promise of data-driven AI models is discovering patterns beyond human reach.

To that end, while TRUST’s empirical foundations center on human-machine partnerships, the AI landscape compels us to envision expanding TRUST to the emerging paradigm of autonomous multi-agent systems. This prospective adaptation requires a deliberate expansion of core terminology: stakeholders broaden to incorporate machine orchestrators, domain experts will include specialized critic agents. Correspondingly, the evaluation of an interpretation method will extend to the evaluation of inter-agent telemetry, or communication protocols between agents in a multi-agent network. Notably, despite contestability being an inherently human-centric concept, its core mechanisms of transparency, interrogation and steerability are equally critical in this agentic paradigm for machine orchestrators to detect misalignment and prevent silent failure cascades.

Regardless of whether an interpretation method bridges AI to human intuition or to automated AI auditors, when algorithmic systems are deployed in a society they become sociotechnical and demand contestability. AI systems are meant to augment or replace existing entities, whether they be established processes, incumbent domain experts or liable gatekeepers. From a practical point of view, interpretation methods are an important path for the incumbent to vet the soon-to-be-deployed AI system. The tiered TRUST framework becomes indispensable: not all interpretation methods are equally trustworthy for deployment readiness and automated agentic systems require proven telemetry fidelity. Interpretation techniques that satisfy all five tiers of TRUST establish the necessary foundation for deploying, monitoring, and contesting AI systems safely. As KDD practices and the broader AI deployment landscape continue to evolve, TRUST provides a flexible scaffold that can incorporate additional tiers to preserve domain-grounded knowledge discovery and accountable, contestable governance.

\begin{acks}
This work was supported by the Research Council of Norway, TRUST-The Norwegian Centre for Trustworthy AI, project number 357425. We also gratefully acknowledge partial support from NSF grant DMS-2413265, NSF grant DMS 2209975, NSF grant DMS-2515767, NSF grant 2023505 on Collaborative Research: Foundations of Data Science Institute (FODSI), the NSF and the Simons Foundation for the Collaboration on the Theoretical Foundations of Deep Learning through awards DMS-2031883 and 814639, and NSF grant MC2378 to the Institute for Artificial CyberThreat Intelligence and OperatioN (ACTION).
\end{acks}

\bibliographystyle{ACM-Reference-Format}
\bibliography{references}

\end{document}